\documentclass{article}

\usepackage{graphicx}
\usepackage{amsmath}
\usepackage{bm}
\usepackage{amsfonts}
\usepackage{amssymb}
\usepackage{algorithm}
\usepackage{algpseudocode}
\usepackage{book tabs}
\usepackage{color}
\usepackage{caption}
\usepackage{subcaption}
\usepackage{rotating}
\usepackage[margin=1.2in]{geometry}
\usepackage{cite}
\usepackage{authblk}
\usepackage{textcomp}
\usepackage{flafter}
\usepackage{hyperref}
\usepackage{cleveref,color}
\usepackage[numbers, sort&compress]{natbib}

\usepackage[table,x11names, svgnames, rgb]{xcolor}
\usepackage[utf8]{inputenc}
\usepackage{tikz}
\usetikzlibrary{snakes,arrows,shapes}

\newcommand{\fref}[1]{Figure~\ref{#1}}
\newcommand{\frefs}[2]{Figures~\ref{#1} and \ref{#2}}

\newcommand{\sref}[1]{Section~\ref{#1}}
\newcommand{\eref}[1]{Equation~\eqref{#1}}
\newcommand{\tref}[1]{Table~\ref{#1}}

\newcommand{\algoref}[1]{Algorithm~\ref{#1}}

\newcommand{\PreserveBackslash}[1]{\let\temp=\\#1\let\\=\temp}
\newcolumntype{C}[1]{>{\PreserveBackslash\centering}p{#1}}
\newcolumntype{R}[1]{>{\PreserveBackslash\raggedleft}p{#1}}

\title{Automated Learning of Interpretable Models with Quantified Uncertainty}

\author{G.F. Bomarito$^*$}
\author{P.E. Leser$^*$}
\affil{National Aeronautics and Space Administration, Langley Research Center, Hampton, VA}
\author{N.C.M. Strauss}
\author{K.M. Garbrecht}
\author{J.D. Hochhalter}
\affil{University of Utah, Salt Lake City, UT}

\begin{document}

\maketitle
\def\thefootnote{*}\footnotetext{These authors contributed equally to this work}\def\thefootnote{\arabic{footnote}}

\begin{abstract}

Interpretability and uncertainty quantification in machine learning can provide justification for decisions, promote scientific discovery and lead to a better understanding of model behavior. Symbolic regression provides inherently interpretable machine learning, but relatively little work has focused on the use of symbolic regression on noisy data and the accompanying necessity to quantify uncertainty. A new Bayesian framework for genetic-programming-based symbolic regression (GPSR) is introduced that uses model evidence (i.e., marginal likelihood) to formulate replacement probability during the selection phase of evolution. Model parameter uncertainty is automatically quantified, enabling probabilistic predictions with each equation produced by the GPSR algorithm. Model evidence is also quantified in this process, and its use is shown to increase interpretability, improve robustness to noise, and reduce overfitting when compared to a conventional GPSR implementation on both numerical and physical experiments.

\end{abstract}

%
%

\section{Introduction}
Machine learning (ML) has become ubiquitous in scientific disciplines.  In some applications, accurate data-driven predictions are all that is required; however, in many others, interpretability and explainability of the model is equally important. Interpretability and explainability can provide justification for decisions, promote scientific discovery and ultimately lead to better control/improvement of models \cite{adadi2018peeking, du2019techniques}.  In a complementary fashion, ML models can provide further insight by conveying their level of uncertainty in predictions \cite{bhatt2020explainable}.  Especially in cases of low risk tolerance this type of insight is crucial for building trust in ML models \cite{rudin1811please}.

Rather than focus on black-box ML methods (e.g., neural networks or Gaussian process regression) combined with post hoc explainability tools, the current work focuses on inherently interpretable methods.  Interpretable ML methods can be competitive with black-box ML in terms of accuracy and do not require a separate explainability toolkit \cite{rudin1811please, rudin2019we}.  Symbolic regression is one such inherently interpretable form of ML wherein an analytic equation is produced that best models input data.  Symbolic regression has been successful in a range of scientific applications such as deriving conservation laws in physics \cite{schmidt2009distilling}, inferring dynamic relationships \cite{brunton2016discovering, galioto2020bayesian}, and producing interpretable mechanics models \cite{bomarito2021development}.  Unfortunately, little attention has been paid to the use of symbolic regression on noisy data and the consideration of uncertainty.  

\citet{schmidt2007learning} tackled the problem of noisy training data in symbolic regression through the inclusion of uniform random variables in model formation.  Though uniform random variables can be transformed to represent more complex distributions, doing so drastically increases complexity of equations that must be produced.  This can make the symbolic regression process less tractable and less interpretable.  

\citet{hirsh2021sparsifying} incorporated Bayesian uncertainty quantification into the sparse identification of nonlinear dynamics (SINDy) method through the use of sparsifying priors.  In this technique, a linear combination of candidate terms (i.e., simple functions of the input data) is produced with random coefficients that are estimated through Bayesian inference.  The reliance on candidate terms and linear combinations thereof constitutes only a limited form of symbolic regression (as opposed to the more traditional free-form symbolic regression).  As such, the form of the resulting equation may be overly constrained and less insightful.

Others have implemented Bayesian methods in symbolic regression \cite[e.g.,][]{jin2019bayesian, zhang2000bayesian}; however, they focused more on the improved efficiency of symbolic regression methods rather than the ability to produce probabilistic models with quantified uncertainty.  For instance, \citet{jin2019bayesian} used a form of Markov chain Monte Carlo as a means for equation production.  Also,  \citet{zhang2000bayesian} used a Bayesian framework to influence the population dynamics in genetic programming for improved evolution speed and decreased complexity.

In the current work, a new Bayesian framework for genetic-programming-based symbolic regression (GPSR) is developed.  In this framework, Baysian inference is applied to infer unknown distributions of parameters in free-form equations.  The marginal likelihood of the equations are then used in a Bayesian model selection scheme to influence evolution towards equations for which the data provides the most evidence.  The result is a GPSR framework that can produce interpretable models with quantified uncertainty.  Additionally, the Bayesian framework provides regularization with several benefits compared to standard GPSR: increased interpretability, increased robustness to noise, and less tendency to overfit.

%
%
%
%
%
%
%
%
%
%
%
%
%
%

\section{Methods}
Symbolic regression is the search for analytic equations that best describe some dataset: i.e., attempting to find a function $f: \mathbb{R}^d \rightarrow \mathbb{R}$ such that $f(\mathbf{x})=y$ given a dataset $\mathcal{D}\{(\mathbf{x}_i,y_i)\}^N_{i=0}$ with $d$-dimensional input features $\mathbf{x}$ and label $y$. Several methodologies have been applied to the task of free-form symbolic regression such as genetic programming \cite{koza1992genetic}, prioritized grammar enumeration \cite{worm2013prioritized}, Markov chain Monte Carlo \cite{jin2019bayesian}, divide and conquer \cite{udrescu2020ai}, and deep learning \cite{petersen2019deep, valipour2021symbolicgpt}.  Genetic programming-based symbolic regression (GPSR) is perhaps the most popular and successful \cite{la2021contemporary}.  The focus of the current work is the integration of uncertainty quantification into the GPSR framework for the robust selection of models when  data is noisy.  This section will first outline a conventional GPSR framework, then describe how it can be modified for consideration of uncertainty.

\subsection{Conventional GPSR implementation}
\label{sec:basicGPSR}
GPSR is an evolutionary approach to symbolic regression, wherein a population of candidate equations is evolved until they adequately describe the input dataset.  The population is first randomly initialized; then, in subsequent generations, the population is evolved through crossover and mutation before entering a selection phase.  The open-source Python package \texttt{Bingo}\cite{bingo} is used in this work for GPSR.

\begin{figure}
  \centering
  \begin{tikzpicture}[>=latex,line join=bevel,scale=0.6]
  \pgfsetlinewidth{1bp}
\pgfsetcolor{black}
  \draw [->] (69.52bp,73.662bp) .. controls (62.488bp,64.456bp) and (53.565bp,52.776bp)  .. (39.431bp,34.273bp);
  \draw [->] (86.202bp,72.202bp) .. controls (88.082bp,64.241bp) and (90.342bp,54.667bp)  .. (94.787bp,35.843bp);
  \draw [->] (60.778bp,144.57bp) .. controls (63.987bp,136.32bp) and (67.894bp,126.27bp)  .. (75.187bp,107.52bp);
  \draw [->] (50.635bp,144.05bp) .. controls (46.027bp,119.48bp) and (37.706bp,75.101bp)  .. (30.407bp,36.173bp);
\begin{scope}
  \definecolor{strokecol}{rgb}{0.0,0.0,0.0};
  \pgfsetstrokecolor{strokecol}
  \draw (82.0bp,90.0bp) ellipse (27.0bp and 18.0bp);
  \draw (82.0bp,90.0bp) node {$\times$};
\end{scope}
\begin{scope}
  \definecolor{strokecol}{rgb}{0.0,0.0,0.0};
  \pgfsetstrokecolor{strokecol}
  \draw (27.0bp,18.0bp) ellipse (27.0bp and 18.0bp);
  \draw (27.0bp,18.0bp) node {$x_0$};
\end{scope}
\begin{scope}
  \definecolor{strokecol}{rgb}{0.0,0.0,0.0};
  \pgfsetstrokecolor{strokecol}
  \draw (99.0bp,18.0bp) ellipse (27.0bp and 18.0bp);
  \draw (99.0bp,18.0bp) node {$x_1$};
\end{scope}
\begin{scope}
  \definecolor{strokecol}{rgb}{0.0,0.0,0.0};
  \pgfsetstrokecolor{strokecol}
  \draw (54.0bp,162.0bp) ellipse (27.0bp and 18.0bp);
  \draw (54.0bp,162.0bp) node {$+$};
\end{scope}
\end{tikzpicture}
  \caption{Acyclic graph representation of the equation $f(\mathbf{x})=x_0+x_0x_1$ with complexity of 4.}
  \label{fig:agraph_equation}
\end{figure}
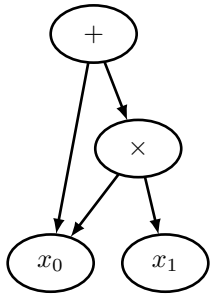

The internal representation of equations plays an important role in GPSR \cite{schmidt2007comparison}.  For example, equations can be represented as an acyclic graph (see \fref{fig:agraph_equation}), where the number of nodes in the acyclic graph is a measure of the equation complexity.  A common difficulty in GPSR is equation bloat (i.e., the tendency to produce increasingly complex equations), which counteracts interpretability. The acyclic graph encoding is chosen in this work, rather than the more common tree encoding, for its superior computational performance and reduced bloat \cite{schmidt2007comparison}.  

The random initialization, crossover, and mutation of equations is based on generating, mixing, and modifying acylic graph components.  A brief description of these algorithms is outlined here; for a more complete description see the \texttt{Bingo} repository \cite{bingo}. Random initialization of an equation consists of generating a sequence of nodes and directed connections.  Single point crossover is used to swap sections of two parent equations resulting in two new children.  Mutation occurs as a random choice from the following: point (node) mutation; edge (directed connection) mutation; node and edge mutation; prune mutation; and branch mutation.  Notably, all of these operations preserve the maximum acyclic graph size and as such a maximum complexity limit can be set.

The consideration of real-valued numerical constants has been a difficulty for GPSR in the past.  Existence of these parameters in equations can increase the search domain and increase bloat pressure.  In early attempts, these parameters would be represented as fixed-value terminals that then must be combined or mutated in equations to allow for derived quantities.  This placed undue burden upon the evolutionary optimization and long periods of evolution could be devoted to simple variation of numerical parameters.  More recently there has been a push for the inclusion of local optimization \cite{kommenda2013effects, de2015evaluating} of these parameters.  In this approach, real-valued numerical constants are treated as abstract placeholders: i.e., equations $f^*(\mathbf{x}, \boldsymbol{\theta})$ include an explicit dependence upon numerical parameters $\boldsymbol{\theta}\in \mathbb{R}^p$.  These parameters are then optimized based on the training data $\boldsymbol{\theta}^* = \mathrm{argmin}_{\boldsymbol{\theta}} ~R(f^*, \mathcal{D}, \boldsymbol{\theta})$  to yield a final equation $f(\mathbf{x}) = f^*(\mathbf{x}, \boldsymbol{\theta}^*)$; here $R$ is an error metric such as root mean squared error (RMSE).  In this way, the optimization of numerical constants is separated from the evolution of the equation form.  Local optimization of numerical constants with RMSE is used in the current work for the conventional GPSR implementation.  As will be shown in \sref{sec:bayesian_methods}, the optimization of constants in this manner allows for a natural probabilistic extension wherein the constants can be considered random variables and the optimization is used for initializing their estimation.

The last remaining aspect of the GPSR implementation in this work is the choice of selection method. The selection process is an important factor in the evolutionary dynamics of the population. The most common form of selection in genetic programming is tournament selection, where individuals in the population compete against other randomly selected individuals for entrance into the next generation, and the individual with best fitness survives. Commonly, an error metric such as RMSE is used to measure fitness in symbolic regression.  Newer selection algorithms have been aimed at reducing the tendency to prematurely converge to a local optimum (i.e., combating the tendency to exploit rather than explore) \cite[e.g.,][]{mahfoud1995niching,schmidt2011age}.  The conventional GPSR implementation in the current work uses one of these forms of selection known as deterministic crowding \cite{mahfoud1995niching}.  In deterministic crowding, individuals are paired with their most-similar offspring (pairing phase); the individual with better fitness survives into the next generation (replacement phase).  Probabilistic extensions have been made to the replacement phase of the deterministic crowding method \cite{mengshoel1999probabilistic, galan2010generalized}; however, the form of the replacement probability has been completely heuristic.  \sref{sec:bayesian_methods} will illustrate how a probabilistic formulation for equation fitness can allow for the more principled use of Bayesian model selection in replacement.

\subsection{Bayesian GPSR implementation}
\label{sec:bayesian_methods}
Extending GPSR to the quantification of uncertainty requires a probabilistic reformulation of the local optimization. Here, $\boldsymbol{\theta}\in \mathbb{R}^p$ is now considered a realization of a random variable $\boldsymbol{\Theta}$ such that $\boldsymbol{\theta} = \boldsymbol{\Theta}(\omega)$, where $\omega \in \Omega$ is a random event in the sample space $\Omega$. The solution to the inverse problem is the probability distribution of $\boldsymbol{\Theta}$ conditioned on both the data, $\mathcal{D}$, and a particular GPSR-generated model, $f$, referred to as the posterior distribution. Note that the explicit dependence of $f(\mathbf{x}, \boldsymbol{\theta})$ on $\mathbf{x}$ is dropped from the notation in the following discussion to preserve clarity. The posterior probability density function (PDF) is given by Bayes' Theorem
\begin{equation}\label{eq:bayes}
	\pi(\boldsymbol{\theta}|\mathcal{D},f) = \frac{\pi(\mathcal{D}|\boldsymbol{\theta},f)\pi(\boldsymbol{\theta}|f)}{\pi(\mathcal{D}|f)}= \frac{\pi(\mathcal{D}|\boldsymbol{\theta},f)\pi(\boldsymbol{\theta}|f)}{\int_{\mathbb{R}^p}\pi(\mathcal{D}|\boldsymbol{\theta},f)\pi(\boldsymbol{\theta}|f)d\boldsymbol{\theta}},
\end{equation}
where $\pi(\boldsymbol{\theta}|f)$ is the prior density function, which encodes \emph{a priori} knowledge about the probability of the parameters, and $\pi(\mathcal{D}|\boldsymbol{\theta}, f)$ is the likelihood function, which represents the probability density for observation $\mathcal{D}$ when $\boldsymbol{\theta}$ takes a specific value.

The form of the likelihood function is dependent on the relationship between $f$ and $\mathcal{D}$. A typical assumption used in Bayesian inference is that there is some noise, $\varepsilon$, associated with the measurements such that
\begin{equation}\label{eq:stochmodel}
	\mathcal{D} = f(\boldsymbol{\theta}) + \varepsilon,
\end{equation}
where the noise is independently and identically distributed according to a zero-mean Gaussian distribution, $N(0, \sigma^2)$, with unknown variance $\sigma^2$. The likelihood function associated with \eref{eq:stochmodel} is known in closed form. This form can be modified depending on the nature of the problem (e.g., the noise could be multiplicative or follow some other distribution) and the likelihood function would change accordingly. \eref{eq:stochmodel} is used for the examples presented herein.

The denominator $\pi(\mathcal{D}|f)$ in \eref{eq:bayes} is referred to as the marginal likelihood and is a normalizing constant to ensure the posterior PDF is proper (i.e., integrates to 1) for a given model. This potentially high-dimensional integral is typically ignored since it is difficult to compute and not required for random walk algorithms such as Markov chain Monte Carlo (MCMC), which exploit the proportionality $\pi(\boldsymbol{\theta}|\mathcal{D},f) \propto \pi(\mathcal{D}|\boldsymbol{\theta},f)\pi(\boldsymbol{\theta}|f)$ to draw samples from the posterior distribution. However, the marginal likelihood is a powerful quantity in Bayesian model selection as it can be used to compute the relative posterior probability of one model versus another using Bayes' factor (BF),
\begin{equation}\label{eq:basicbf}
	B = \frac{\pi(f_1|\mathcal{D})}{\pi(f_2|\mathcal{D})}=\frac{\pi(\mathcal{D}|f_1)\pi(f_1)}{\pi(\mathcal{D}|f_2)\pi(f_2)},
\end{equation}
where $\pi(\mathcal{D}|f_i)$ and $\pi(f_i)$ are the marginal likelihood and prior probability of the $i^{\mathrm{th}}$ model, respectively. In the GPSR context, the BF can be used to measure relative fitness when comparing equations during the selection phase. This has the dual advantage of accounting for uncertainty and penalizing complexity through what has been referred to as the Bayesian Occam's Razor \cite{murray2005note}.  The marginal likelihood $\pi(\mathcal{D}|f)$ represents the space of possible data that can be represented by a given model. As the dimension $p$ increases and the model becomes more flexible, the probability density function of the data is dispersed.


Judicious selection of prior distributions is a critical part of Bayesian inference and particularly model selection due to the sensitivity of $\pi(\mathcal{D}|f)$ to both prior bounds and dimensionality. Subjective versus objective priors is a point of contention in Bayesian statistics \cite{goldstein2006subjective, berger2006objective}. In the context of GPSR, where models are being generated from an immense space of operators and combinations thereof, it would be challenging if not impossible to be subjective; e.g., to have an expert perform elicitation or to use previous experimentation to influence the selection of priors for an unknown model. Priors for the Bayesian GPSR implementation are thus aimed at being objective in that the inference is only influenced by data to the extent possible. For example, $\pi(f_i)=\pi(f_j)$ was assumed here to not favor any particular model over another, yielding a BF equal to the ratio of marginal likelihoods.

Objective choices for $\pi(\boldsymbol{\theta}|f)$ are more challenging as optimally non-informative priors depend on the form of the model \cite{berger2015overall}. A simple choice, although suboptimal in terms of reducing the influence of the prior, is an improper uniform distribution $\pi(\boldsymbol{\theta}|f)\propto1$ for $\boldsymbol{\theta} \in \mathbb{R}^p$. However, this causes issues when computing BF due to indeterminate constants appearing in \eref{eq:basicbf}. O'Hagan introduced the fractional Bayes factor (FBF) \cite{ohagan1995fbf} to address this issue:
\begin{equation}
	B_\gamma = \frac{q_1(\gamma)}{q_2(\gamma)},
\end{equation}
where
\begin{equation}
	q_i(\gamma) = \frac{\int_{\mathbb{R}^p}\pi(\mathcal{D}|\boldsymbol{\theta},f_i)\pi(\boldsymbol{\theta}|f_i)d\boldsymbol{\theta}}{\int_{\mathbb{R}^p}\pi(\mathcal{D}|\boldsymbol{\theta},f_i)^\gamma\pi(\boldsymbol{\theta}|f_i)d\boldsymbol{\theta}}.
\end{equation}
The unknown constants in the marginal likelihood are normalized out using the $\gamma \in (0, 1]$ power of the likelihood function, allowing for consistent model selection when using improper uniform priors to compute BF. Following O'Hagan's recommendation for improved robustness to prior misspecification, $\gamma=1\sqrt{N}$ was chosen for this work.

Estimating marginal likelihood is challenging using MCMC, and computing the FBF requires two separate runs of an MCMC sampler to target the additional normalizing $\gamma$ posterior. Sequential Monte Carlo (SMC) evolves a set of weighted particles through a sequence of reweighting, resampling, and local abbreviated MCMC steps and can be used as a drop-in replacement for MCMC \cite{del2006sequential}. SMC has several advantages over MCMC, including the ability to produce direct, unbiased estimates of the marginal likelihood, $\pi(\mathcal{D}|f)$, as well as potential reductions in computation time relative to MCMC through parallelization of the $\pi(\mathcal{D}|\boldsymbol{\theta},f)$ evaluations. Furthermore, SMC relies on a sequence of target distributions based on an annealed likelihood, which is conveniently the same form as the FBF; i.e., $\pi(\mathcal{D}|\boldsymbol{\theta},f)^{\phi_t}$ with $\phi_t$ monotonically increasing from 0 at $t=0$ to 1 at $t=1$ during the sequential sampling process. Therefore, $q_i(\gamma)$ is a natural byproduct of a single run of an SMC sampler if $\phi_t=\gamma$ is included as the $t^\mathrm{th}$ step of the sequence and the marginal likelihood is estimated at both the $t^\mathrm{th}$ and final steps.

Initialization of the SMC algorithm is a practical challenge when considering arbitrary equations with priors over the entire real space $\mathbb{R}^p$. To prevent early divergence and degeneracy of the particles (i.e., a significant portion of particles having zero weight), it is important to locate the initial particles near to and encompassing the unknown region of non-zero posterior probability. A multistart local optimization approach is adopted here to quickly identify local RMSE minima. These local minima are considered potential regions of high posterior probability. Multivariate normal distributions (MVN) centered at each local minimum are used to approximate the posterior distribution,
\begin{equation}
	\pi(\boldsymbol{\theta}|\mathcal{D},f) \approx MVN(\boldsymbol{\theta}^*, \Sigma^*),
\end{equation}
where the availability of partial derivatives is exploited to estimate the covariance matrix. The restriction of $N > p$ was relaxed in the symbolic regression framework so, in place of the unbiased estimator in \cite{smith2013uncertainty} with denominator $N-p$, a biased\footnote{Errors associated with this approximation only influence the inference process through the placement of initial particles, and the effect was assumed to be minimal.} estimator was used,
\begin{align}
	\Sigma^* &= \frac{\sum_{i=0}^N [y_i - f(x_i, \boldsymbol{\theta}^*)]^2}{N} (\chi^T\chi)^{-1},\\
	\chi_{i,k} &= \frac{\partial f(x_i, \boldsymbol{\theta}^*)}{\partial \theta_k^*}.
\end{align}

The SMC algorithm is initialized by sampling particles from each MVN distribution in equal proportion (i.e., there was no attempt to combine co-located distributions). The number of multistarts dictates how much the initial population is allowed to explore the infinite parameter space, with an increasing number of multistarts corresponding to increasing the chance of SMC identifying all modes of the true posterior distribution.

The proposed Bayesian GPSR implementation is an extension of probabilistic crowding \cite{mengshoel1999probabilistic}, where the fitness and thus the replacement probability is based on the FBF. The open-source, vectorized SMC Python implementation \texttt{SMCPy} \cite{smcpy} is used to efficiently compute normalized marginal likelihoods $q_i(\gamma)$ for each individual. After pairing most-similar offspring, a replacement probability is defined as $p(f_1|q_1,q_2)=q_1/(q_1+q_2)$. For FBF$=1$, this selection probability is equal to 0.5. For cases where FBF $> 1$, the first model is more likely to proceed to the next generation, etc. See \algoref{algo:smcbingo} for a summary.

\begin{algorithm}
\caption{Probabilistic Crowding using Bayesian Model Selection}
\label{algo:smcbingo}
\begin{algorithmic}[1]
\Require Population of equations at $i^\mathrm{th}$ generation ($\boldsymbol{F}_i$), training data ($\mathcal{D}$)
\For{most-similar pairs $(f_1, f_2)$ in $\boldsymbol{F}_i$}
	\State $q_1 \gets \mathrm{SMC}(f_1, \mathcal{D})$
	\State $q_2 \gets \mathrm{SMC}(f_2, \mathcal{D})$
	\State $p(f_1|q_1, q_2) = q_1 / (q_1 + q_2)$
	\State Sample $u \sim \textrm{Uniform}(0, 1)$
	\If{$u \leq p(f_1|q_1,q_2)$}
		\State $\boldsymbol{F}_{i+1} \gets f_1$	
	\Else
		\State $\boldsymbol{F}_{i+1} \gets f_2$
	\EndIf
\EndFor
\end{algorithmic}
\end{algorithm}

\subsection{Demonstration}
\begin{figure}
  \centering
  \begin{subfigure}[b]{0.34\textwidth}
         \centering
         \includegraphics{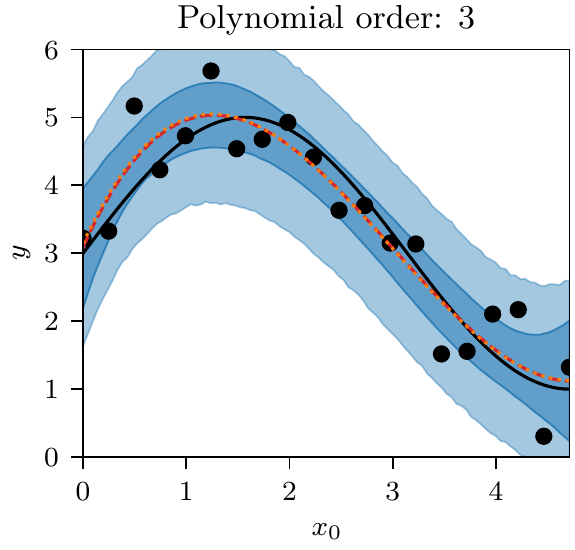}
     \end{subfigure}
     \hfill
     \begin{subfigure}[b]{0.62\textwidth}
         \centering
         \includegraphics{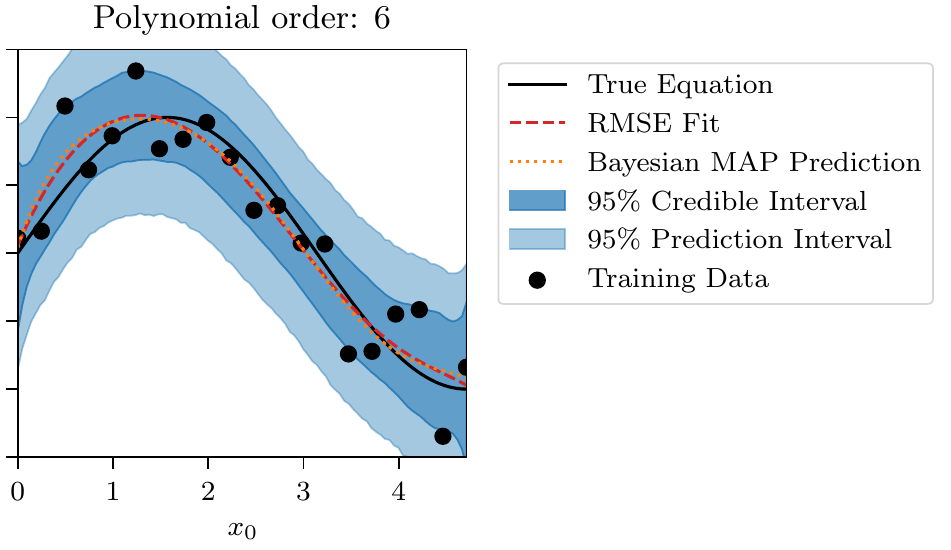}
     \end{subfigure}
  \caption{Polynomial models fit to noisy data using SMC (probabilistic) and RMSE (deterministic). Data was generated from $y=2\sin{x_0}+3$ with added Gaussian noise.  The probabilistic fit is illustrated with the MAP prediction and the 95\% credible/prediction intervals. }
  \label{fig:polynomial_expansion_models}
\end{figure}

To demonstrate the utility of the FBF for GPSR selection as well as the ability of SMC to estimate the FBF, a numerical experiment was conducted. A true function $y=2\sin{x_0}+3$ was defined and synthetic data was generated by adding zero-mean Gaussian noise.\footnote{Note that this example is kept intentionally vague to focus the reader on general fitness metric trends; a more detailed discussion of the example is provided in \sref{sec:numexp}.} Polynomials of increasing order (0 to 6) were fit to the data using two methods: (i) a deterministic minimization of RMSE and (ii) probabilistic parameter estimation using SMC as described in \sref{sec:bayesian_methods}. The deterministic and probabilistic fits for polynomial orders 3 and 6 are compared along with the training data in \fref{fig:polynomial_expansion_models}. Since SMC produces a probabilistic fit of model parameters and simultaneously estimates measurement noise, both the 95\% credible and prediction intervals are shown\footnote{Credible intervals represent model parameter uncertainty while prediction intervals include estimated measurement noise as well.}. The maximum \emph{a posteriori} (MAP) fit, $f(x_0, \boldsymbol{\theta}_\mathrm{MAP})$, where $\boldsymbol{\theta}_\mathrm{MAP}$ are the most probable parameters, is also shown. While the RMSE and MAP fit are very similar as expected,\footnote{Maximum likelihood estimation and MAP are equivalent when using uniform priors on $\boldsymbol{\theta}$, as was the case in this example.} the probabilistic fit expresses increased uncertainty through widening intervals at the edges of the domain where the largest errors with respect to the true model are seen, an advantage over the deterministic fit.

\begin{figure}
  \centering
  \includegraphics{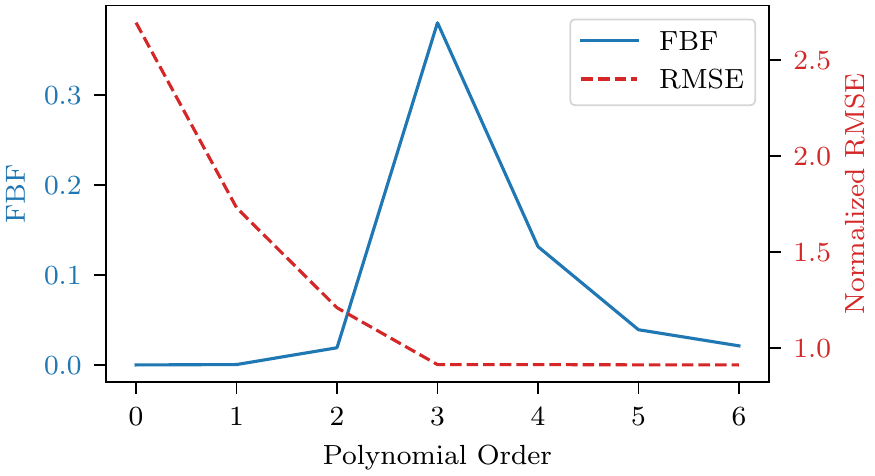}
  \caption{FBF- and RMSE-based fitness for polynomials of increasing order. RMSE is normalized by the standard deviation of the added noise.}
  \label{fig:polynomial_expansion_fitness}
\end{figure}

For each SMC run, the FBF was also computed as an alternative fitness measure to RMSE. Results are shown in \fref{fig:polynomial_expansion_fitness}. As expected, the FBF naturally penalizes complexity as illustrated by a decrease in FBF after polynomial order 3. In contrast, RMSE continues to decrease, showing a propensity for overfitting and equation bloat. Therefore, it is believed that using the FBF fitness metric will lead to identification of more parsimonious equations and improved robustness to noise in the data.


\section{Experiments and Discussion}
In this section, Bayesian GPSR is applied first to a numerical example and then to an experimental example\footnote{Examples with input dimension of $d=1$ are used here for simplicity but Bayesian GPSR can be used in higher dimensions just as other GPSR techniques (see \cite{bomarito2021development} for an example of a similar GPSR implemenation applied to higher-dimensional data).}.  The ability of Bayesian GPSR to produce interpretable models and make probabilistic predictions is illustrated. Comparisons are made to conventional GPSR that illustrate several benefits to the Bayesian extension beyond its ability to produce probabilistic predictions.  

\subsection{Numerical example}\label{sec:numexp}
In a first exposition of Bayesian GPSR, a synthetic example is considered that allows for numerical adjustment and investigation.  In this example, datasets are generated using the function:
\begin{equation}
y = 2\sin(x_0) + 3 + \epsilon,
\label{eq:sinax_plus_b}
\end{equation}
where $\epsilon$ is sampled from a normal distribution with zero mean and standard deviation, $\sigma$.  Here, $\mathbf{x}$ is one-dimensional ($d=1$) and drawn from a uniform distribution with bounds $(0, \frac{3\pi}{2})$.  Several training datasets are generated, each with twenty datapoints ($N=20$), which vary based on $\sigma$ as well as the random seed used for $\mathbf{x}$ and $\epsilon$.  Testing datasets of size $N=1000$ are generated in a similar fashion, but only vary based on $\sigma$ (i.e., a single random seed is used for $\mathbf{x}$ and $\epsilon$).  

For all GPSR runs in this subsection (both conventional and Bayesian) the following hyperparameters are used:  population size of 120, complexity limit of 64\footnote{Equations reaching the maximum complexity of 64 were infrequent; thus, results presented here are expected to be relatively invariant of this choice}, number of generations 1000. The Levenberg-Marquardt method, as implemented in \texttt{SciPy}\cite{2020SciPy-NMeth}, is used for optimization of numerical constants.  GPSR was restricted to evolution of polynomials by limiting the use of mathematical operators to $[+, -, \times]$.  This restriction on operators precludes the ability of finding the true model, allowing for better study of overfitting and generalization properties of the method.

\begin{figure}
  \centering
  \begin{subfigure}[]{0.32\textwidth}
         \centering
         \includegraphics{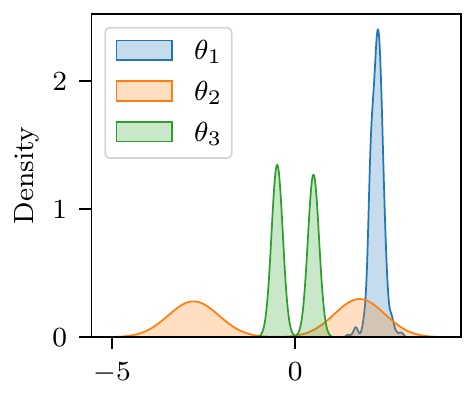}
         \caption{}
  	\label{fig:numerical_param_dists_1}
     \end{subfigure}
  \begin{subfigure}[]{0.32\textwidth}
         \centering
         \includegraphics{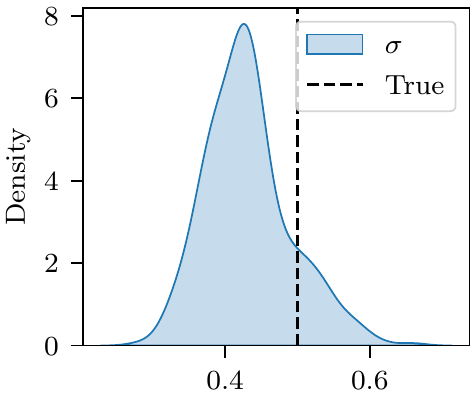}
         \caption{}
  	\label{fig:numerical_param_dists_2}
     \end{subfigure}
  \begin{subfigure}[]{0.32\textwidth}
         \centering
         \includegraphics{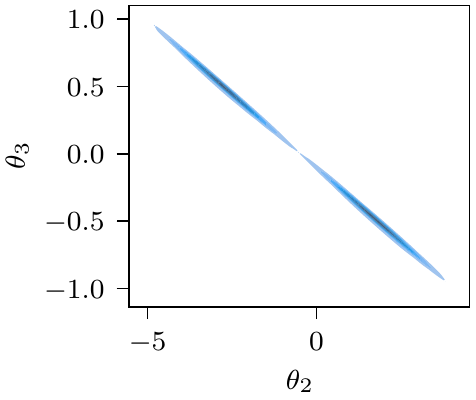}
         \caption{}
  	\label{fig:numerical_param_dists_3}
     \end{subfigure}
  \caption{The distributions accompanying \eref{eq:numerical_example_single_result} as a result of using Bayesian GPSR in the numerical example.  (a) Marginal densities for model parameters, (b) the marginal estimate of the noise scale, and (c) pairwise plot illustrating covariance between two model parameters.  Note that distributions are illustrated with a fitted kernel density estimate as opposed to the discrete particle representation used in SMC.}
  \label{fig:numerical_param_dists}
\end{figure}

For each training dataset, both GPSR and Bayesian GPSR were run once.  The result of one of the Bayesian GPSR runs was the equation
\begin{equation}
f(\mathbf{x}, \boldsymbol{\theta}) = \theta_1 + (\theta_2 + \theta_3 x_0)(x_0 + x_0(\theta_2 + \theta_3 x_0))
\label{eq:numerical_example_single_result}
\end{equation}
with accompanying posterior distributions for $\boldsymbol{\theta}$ and $\sigma$.  The marginal distributions of $\boldsymbol{\theta}$ and $\sigma$ are illustrated in \frefs{fig:numerical_param_dists_1}{fig:numerical_param_dists_2}, respectively.  The $\boldsymbol{\theta}$ parameters are largely uncorrelated to each other except for $\theta_2$ and $\theta_3$ which are highly correlated.  The correlation of these two parameters are illustrated in the pairwise plot in \fref{fig:numerical_param_dists_3}.  It is also seen that $\theta_2$ and $\theta_3$ exhibit bimodal behavior in this result; which indicates the ability of the Bayesian framework to accurately address such scenarios.

\begin{figure}
  \centering
  \includegraphics{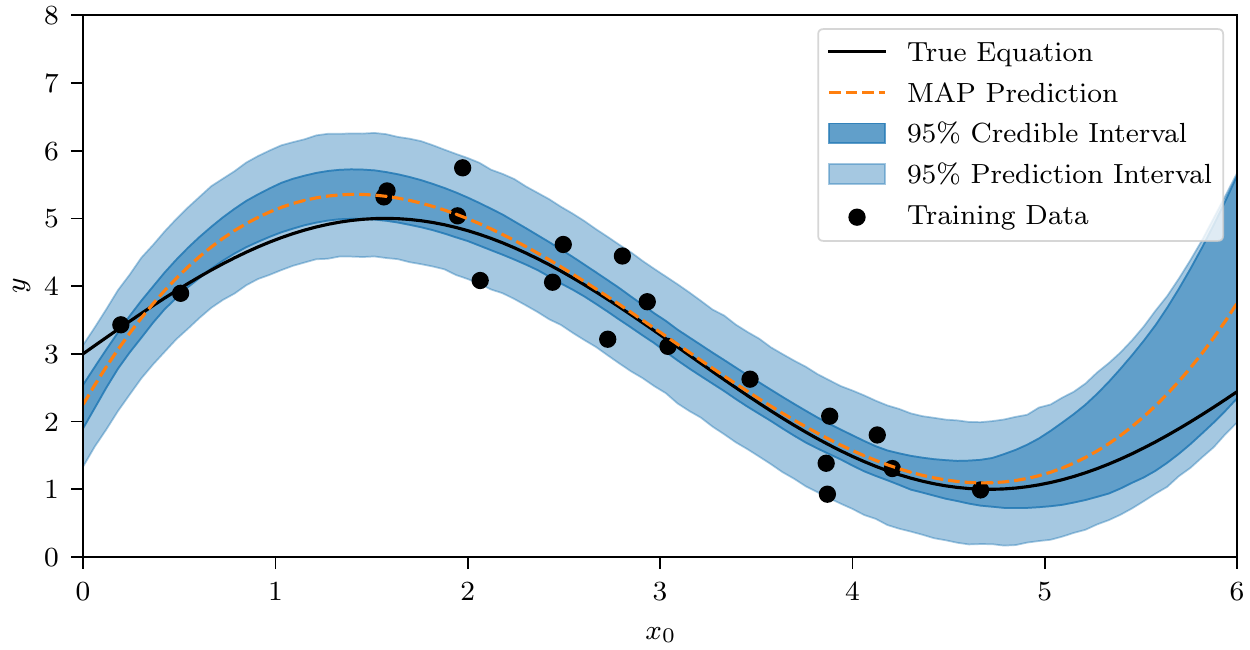}
  \caption{Probabilistic prediction using \eref{eq:numerical_example_single_result} and the SMC-estimated parameter uncertainty compared to the training data and true equation.}
  \label{fig:numerical_prediction}
\end{figure}

The predictive capability of \eref{eq:numerical_example_single_result} is illustrated in \fref{fig:numerical_prediction}.  Most importantly, the model gives probabilistic predictions.  The MAP prediction could be used in cases where a deterministic prediction is needed; credible and prediction intervals can also be used to illustrate the degree of certainty in those predictions.  To help illustrate the importance of the probabilistic prediction, the right side of \fref{fig:numerical_prediction} extends beyond the data and shows the behavior of the model upon extrapolation. There, the degree of uncertainty in the prediction starts to increase, thus indicating that predictions in the extrapolated region are less trustworthy.

\begin{figure}
  \centering
  \begin{subfigure}[b]{0.49\textwidth}
         \centering
         \includegraphics{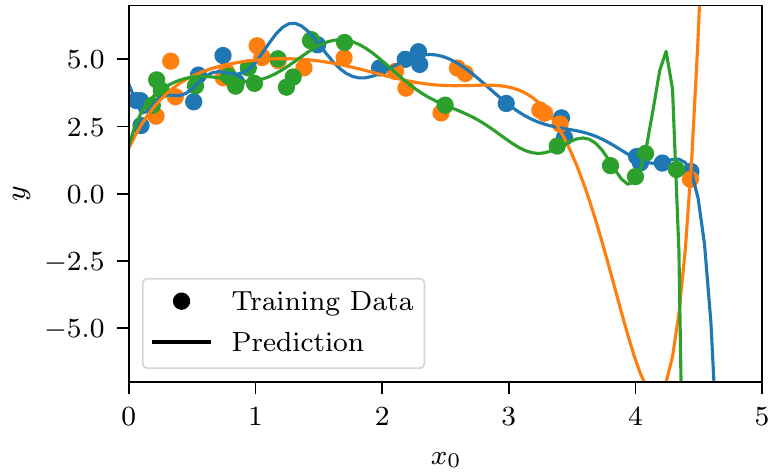}
         \caption{}
     \end{subfigure}
     \hfill
     \begin{subfigure}[b]{0.49\textwidth}
         \centering
         \includegraphics{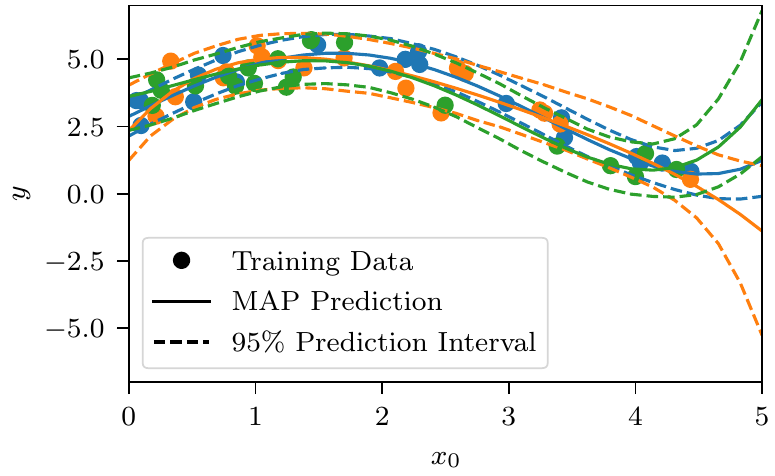}
         \caption{}
     \end{subfigure}
  \caption{Three examples of model predictions produced by (a) conventional GPSR and (b) Bayesian GPSR.}
  \label{fig:sin_3_predictions}
\end{figure}

In addition to having the ability to quantify uncertainty, Bayesian GPSR has several advantages over conventional GPSR.  Firstly, Bayesian GPSR is more robust to noise.  Given the same noisy, data-generating function, Bayesian GPSR produces models which are more invariant to the specifics of the dataset.  \fref{fig:sin_3_predictions} shows the results of Bayesian GPSR and conventional GPSR being trained on the same 3 datasets which were generated using \eref{eq:sinax_plus_b} and different random seeds.  Models produced with conventional GPSR vary wildly based on different datasets whereas Bayesian GPSR produces more noise-invariant models.  The extrapolatory behavior of Bayesian GPSR models does degrade in quality with distance from the training data; however, it represents a large improvement over conventional GPSR.  Conventional GPSR includes no measure of confidence in predictions whereas Bayesian GPSR predictions indicate confidence in its predictions.  As such, users of Bayesian GPSR can know when to exercise caution, such as in extrapolatory regions.

\begin{figure}
  \centering
  \begin{subfigure}[b]{0.49\textwidth}
         \centering
         \includegraphics{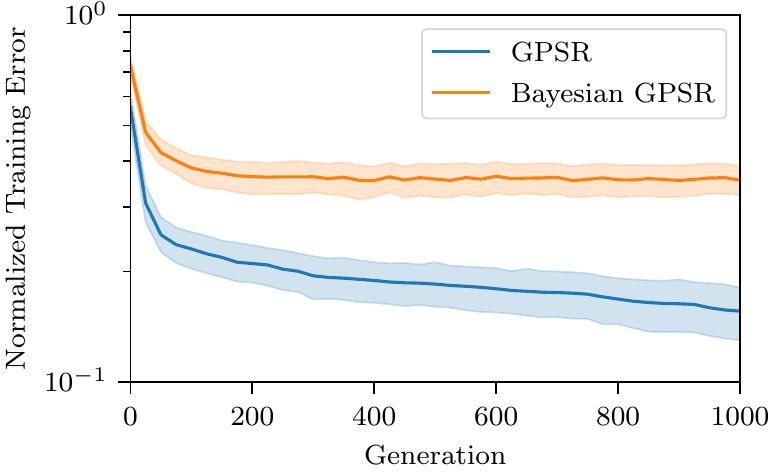}
         \caption{}
     \end{subfigure}
     \hfill
     \begin{subfigure}[b]{0.49\textwidth}
         \centering
         \includegraphics{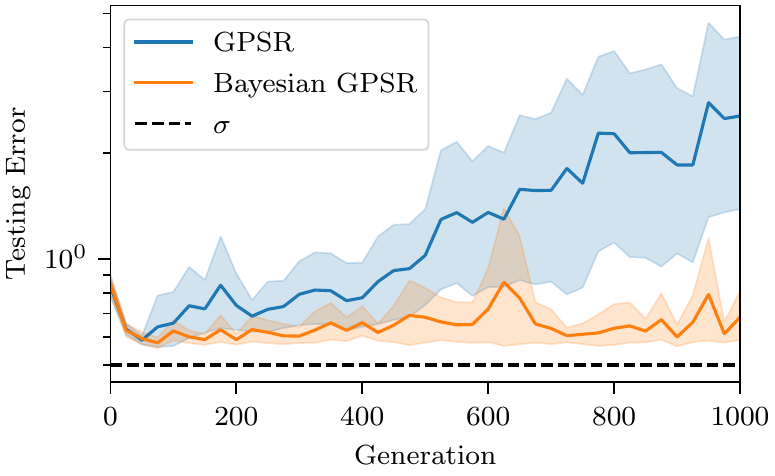}
         \caption{}
     \end{subfigure}
  \caption{Testing and training error for conventional GPSR and Bayesian GPSR.}
  \label{fig:sin_test_train}
\end{figure}

\begin{figure}
  \centering
  \includegraphics[width=0.5\textwidth]{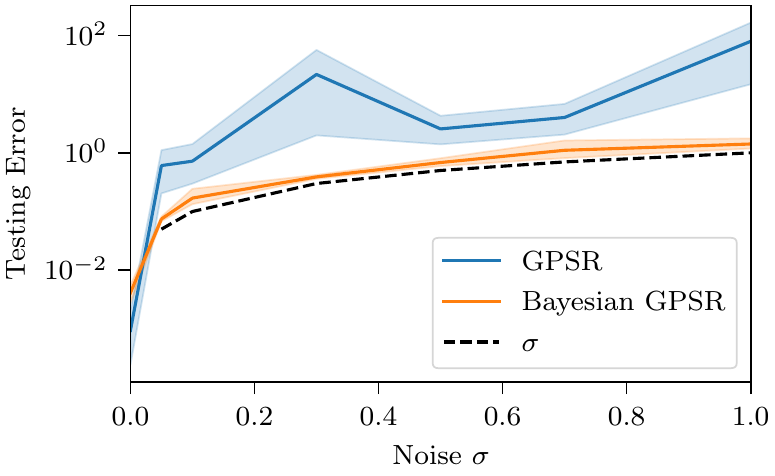}
  \caption{Relationship between testing error and noise level for conventional GPSR and Bayesian GPSR.}
  \label{fig:sin_noise_level_test}
\end{figure}

The use of a Bayesian framework in GPSR provides a level of regularization against overfitting.  This can be seen when comparing the training and test error in \fref{fig:sin_test_train}.  Test error is quantified as the RMSE on the larger test dataset; the true model would have a test error of $\sigma$.  In both Bayesian and conventional GPSR, training error decreases with continued evolution.  In conventional GPSR, however, the test error starts to increase early in the evolution.  The test error for Bayesian GPSR, though oscillating, remains relatively constant and only slightly above the value of $\sigma$.  This trend is consistent across datasets with varying noise levels, as seen in \fref{fig:sin_noise_level_test}.  Bayesian GPSR consistently finds models with test errors near  $\sigma$ and conventional GPSR produces models that are overfit to the training data.  It should be noted that in the zero-noise case, conventional GPSR produces more accurate models. However, once any amount of noise was added to the training data, Bayesian GPSR is preferred.

\begin{figure}
  \centering
  \begin{subfigure}[b]{0.49\textwidth}
         \centering
         \includegraphics[width=\textwidth]{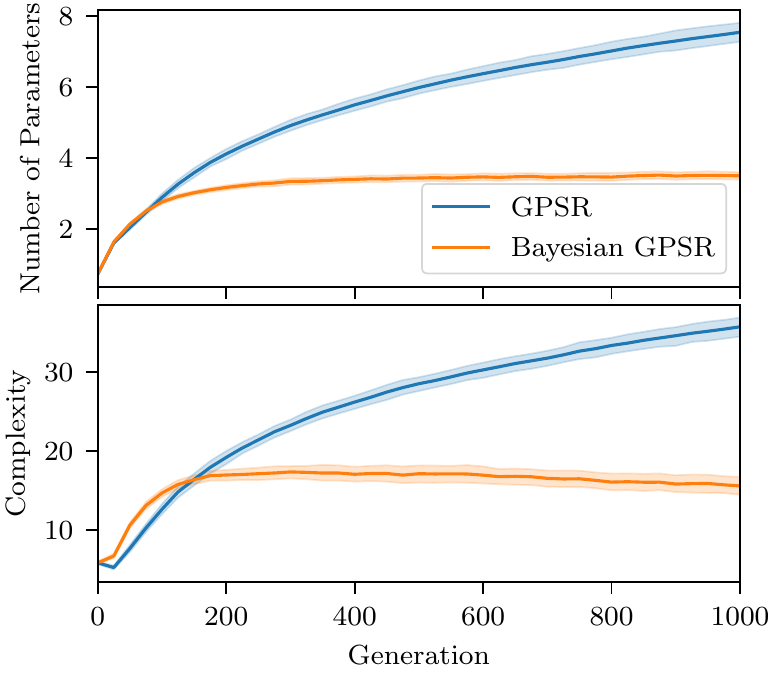}
         \caption{}
  	\label{fig:sin_bloat}
     \end{subfigure}
     \hfill
     \begin{subfigure}[b]{0.49\textwidth}
         \centering
         \includegraphics[width=\textwidth]{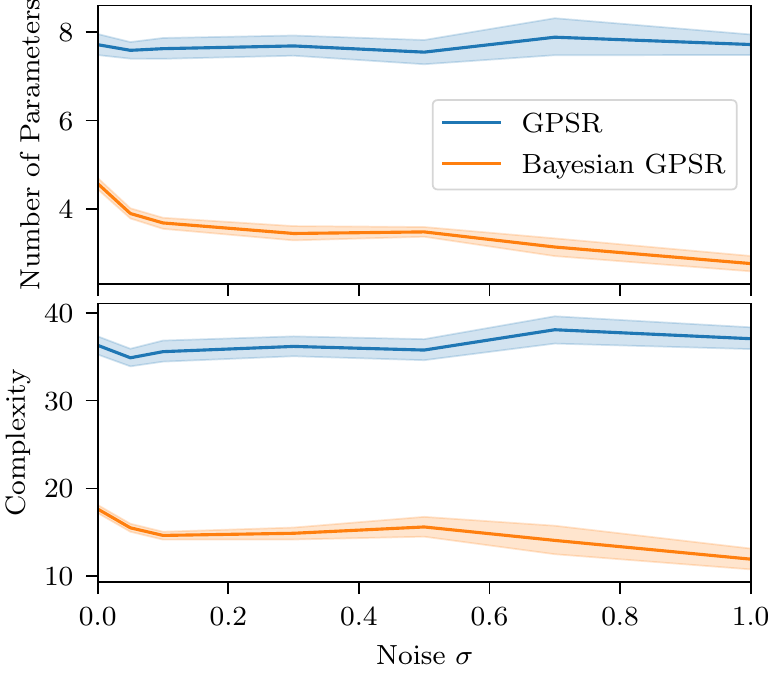}
         \caption{}
 	 \label{fig:sin_noise_level_bloat}
     \end{subfigure}
  \caption{Representations of bloat for conventional GPSR and Bayesian GPSR as a function of (a) generation and (b) level of noise in the data.}
  \label{fig:sin_bloat_full}
\end{figure}

The Bayesian framework also provides GPSR with a level of regularization against the generation of overly complex equations (i.e., bloat).  In conventional GPSR, candidate equations grow continuously in complexity and number of model parameters (i.e., dimension of $\boldsymbol{\theta}$), whereas candidates equations in Bayesian GPSR are less prone to bloat.  Bloat control methods such as parsimony pressure, early stopping, and strict complexity limits (using crossvalidation to select the additional hyperparameters) could be employed in GPSR; however, the objective function in the evolution of the population becomes misaligned with the ultimate model choice.  These methods find a best model encountered in the evolution based on one metric while incentivizing the evolution based on another.  Essentially, these methods become a form of post-hoc model selection rather than a driver of the evolutionary process itself.  \fref{fig:sin_bloat} illustrates that Bayesian GPSR tends to find a complexity that is appropriate for the dataset after which bloat slows or stops.  In contrast to other mechanisms for bloat control, the Bayesian framework automatically identifies an appropriate level of complexity based on the training data.  \fref{fig:sin_noise_level_bloat} shows that in lower noise scenarios Bayesian GPSR permits more complex equations and in high-noise scenarios less complexity is needed.

Lastly, Bayesian GPSR produces models which are significantly more interpretable than conventional GPSR.  This is primarily due to the reduced complexity of resulting equations.\footnote{Neural networks can be represented by an analytic equation, but it is so complex that they are considered black-box.}  For example, a best fit equation of a Bayesian GPSR run with approximately median-complexity is
\begin{equation}
f = \theta_1 + (1 - \theta_2) x_0 + x_0^2 + \theta_2 x_0^3(\theta_3 + \theta_2 x_0),
\end{equation}
whereas a best fit equation of a conventional GPSR run with approximately median complexity is
\begin{equation}
\begin{split}
f = &\theta_1 + x_0 + \theta_2(x_0 + x_0^3(\theta_3 - x_0^2)(\theta_4 + \theta_5 x_0^2(x_0^3 - x_0)) - \theta_5 - x_0^3 - x_0^4) \\
&+ x_0(\theta_6(\theta_7 + x_0^2 + (\theta_8 + x_0^4 -\theta_5)x_0^2(x_0^3 - (x_0)))(\theta_9 + x_0^3 + \theta_{10} x_0^3(x_0^3 - x_0)) \\
& ~~~~~~~~ - (x_0^3(\theta_4 + \theta_5 x_0^2 (x_0^3 - x_0))) - x_0)
\end{split}
\end{equation}
The reduced complexity clearly aids in interpretability.

\subsection{Galileo Example}
In this section, Bayesian GPSR is applied to physical data from the scientific notebooks of Galileo Galilei.  The goal is to illustrate the ability of Bayesian GPSR to develop a predictive and interpretable equation with quantified uncertainty.  The Galileo experiments in this section were chosen specifically because they represent iconic, noisy data in a scenario where the true physics (i.e., data-generating function) of the problem is known.

\begin{figure}
  \centering
  \begin{subfigure}[b]{0.49\textwidth}
         \centering
         \includegraphics[]{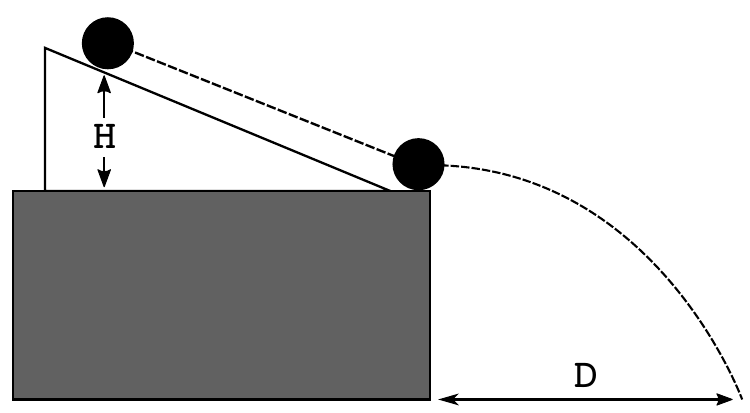}
         \caption{}
  	\label{fig:galileo_diagram_1}
     \end{subfigure}
     \hfill
     \begin{subfigure}[b]{0.49\textwidth}
         \centering
         \includegraphics[]{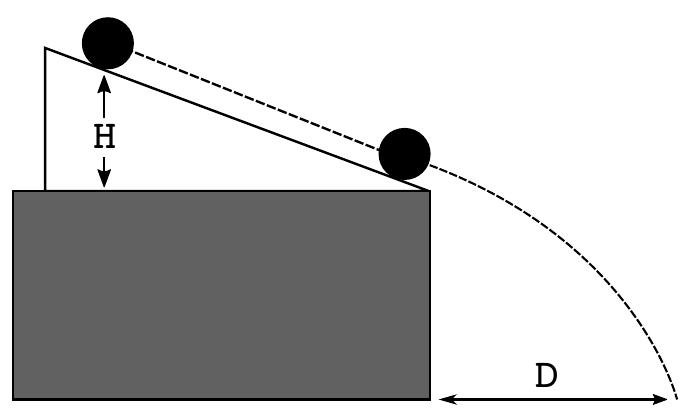}
         \caption{}
  	\label{fig:galileo_diagram_2}
     \end{subfigure}
  \caption{Schematic of Galileo's experiment (a) with shelf and (b) without shelf.}
  \label{fig:galileo_diagram}
\end{figure}

While Galileo was teaching at the University of Padua, he performed a series of experiments studying projectile motion and free fall.  In one of those experiments, he dropped a ball down an inclined plane and then off a horizontal shelf as illustrated in \fref{fig:galileo_diagram_1}.  He recorded the initial drop height of the ball $H$ and the horizontal distance traveled during free fall $D$.  He also performed the same experiment without the horizontal shelf giving an initial downward velocity to the ball (seen in \fref{fig:galileo_diagram_2}).  The data Galileo gathered for these examples\cite{drake2003galileo} are included in \tref{tab:galileo_data}.

\begin{table}
\centering
\caption{Data gathered by Galileo in his free fall experiments \cite{drake2003galileo, dickey1995teaching}. Data is in units of \emph{punti}.}
\label{tab:galileo_data}
\begin{tabular}{cc}
\multicolumn{2}{c}{With Shelf}\\
\hline
$D$& $H$\\
\hline
\hline
         1500 & 1000\\
         1340 & 828\\
         1328 & 800\\
         1172 & 600\\
           800 & 300\\
\hline
\end{tabular}
\quad\quad\quad\quad
\begin{tabular}{cc}
\multicolumn{2}{c}{Without Shelf}\\
\hline
$D$& $H$\\
\hline
\hline
          573 & 1000\\
          534 & 800\\
          495 & 600\\
          451 & 450\\
          395 & 300\\
          337 & 200\\
          253 & 100\\
\hline
\end{tabular}
\end{table}

An identical GPSR setup is used from the previous section, with the exceptions of the mathematical operator set and local optimization algorithm. To enable the discovery of of the true physical equations, an extended set of operators are used: $[+, -, \times, /, \text{pow}, \text{sqrt}]$.  In the Galileo examples, the datasets are much smaller: $N=5$ and $N=7$.  This poses a limitation for Levenberg-Marquardt; i.e., when the number of parameters in a given equation exceeds the number of datapoints.  In these cases, the Broyden–Fletcher–Goldfarb–Shanno (BFGS) algorithm, as implemented in \texttt{SciPy}\cite{2020SciPy-NMeth}, was used for local optimization.

\subsubsection*{Experiment with shelf}

The experiment with the horizontal shelf is the easier of the two to describe mathematically.  The underlying physics is captured by the following relationship \cite{dickey1995teaching}
\begin{equation}
D = k \sqrt{H},
\label{eq:shelf-physics}
\end{equation}
where $k$ is dependent upon the geometry of the table and ramp, and  $k=47.09$ best fits the Galileo dataset in a least-squares sense.

\begin{figure}
\begin{subfigure}[b]{0.49\textwidth}
\includegraphics[]{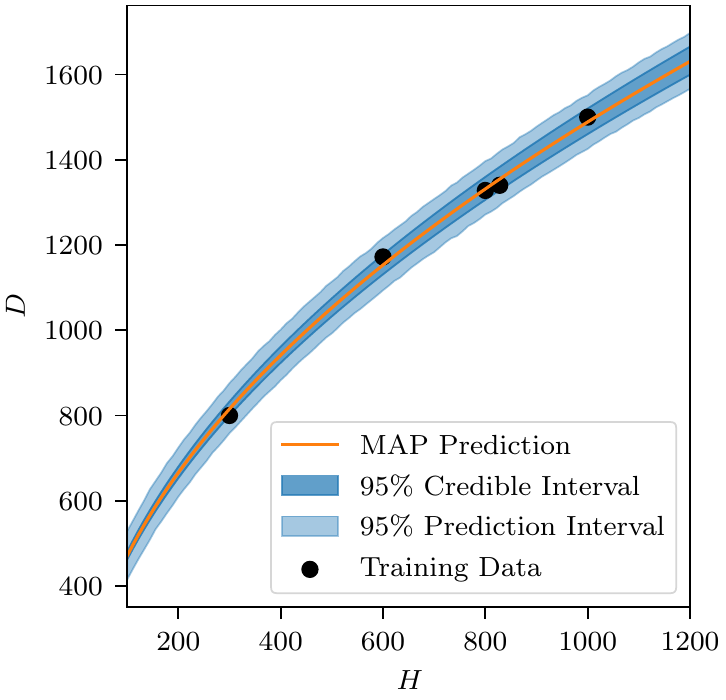}
\caption{}
\end{subfigure}
\hfill        
\begin{minipage}[b]{0.49\textwidth}
\begin{subfigure}[b]{\linewidth}
\includegraphics[]{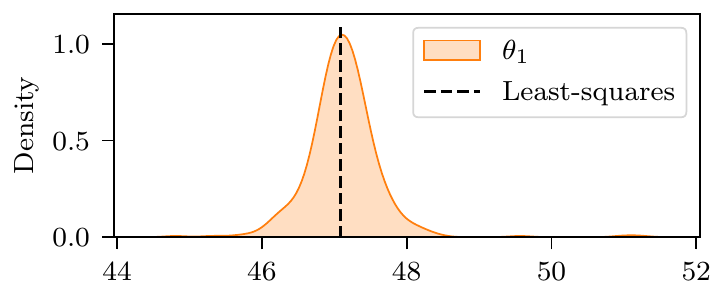}
\caption{}
\label{fig:galileo_shelf_param}
\end{subfigure}

\vspace*{5mm} 
\begin{subfigure}[b]{\linewidth}
\includegraphics[]{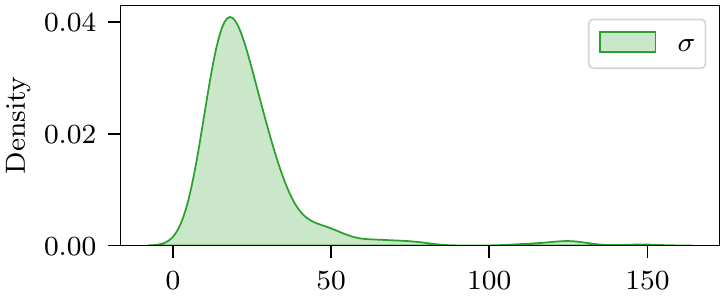}
\caption{}
\label{fig:galileo_shelf_noise}
\end{subfigure}
\end{minipage}
\caption{Representative example of Bayesian GPSR result (\eref{eq:example-shelf}) for the ramp-with-shelf data: (a) probabilistic prediction compared to the training data, (b) estimated probability density function of $\theta_1$, and (c) estimated probability density function of noise standard deviation, $\sigma$.}
\label{fig:galileo_shelf}
\end{figure}

Due to inherent randomness, Bayesian GPSR was repeated five times using Galileo's ramp-with-shelf data.  In all repetitions, the correct model form was found and the final population consisted primarily of different algebraic forms of \eref{eq:shelf-physics}.  A representative result is illustrated in \fref{fig:galileo_shelf}, which has the following form:
\begin{equation}
f(\mathbf{x}, \boldsymbol{\theta}) = \theta_1 \sqrt{x_0}.
\label{eq:example-shelf}
\end{equation}
Not only is the underlying physics identified from this small dataset, but a reasonable description of uncertainty is produced.  The posterior distribution for $\boldsymbol{\theta}$ and $\sigma$ is illustrated in \frefs{fig:galileo_shelf_param}{fig:galileo_shelf_noise}, noting that the MAP $\theta_1$ aligns well with the least-squares fit.

In all of the Bayesian GPSR repetitions, the final population contained numerically approximate equations.  These equations are characterized by an equation form that does not algebraically simplify to the true physics but approximates it numerically.  An example of one such equation is
\begin{equation}
f(\mathbf{x}, \boldsymbol{\theta}) = \frac{\theta_1 \sqrt{x_0}}{1+\theta_2 x_0},
\end{equation}
where $\theta_1$ is distributed similarly to previous equation and $\theta_2 \approx -7.6 \times 10^{-19}$.  Occasionally, these numerically approximate equations are favored over the true physical equation in terms of the FBF, making them overfit solutions to the GPSR problem.  Though the Bayesian GPSR does provide regularization against overfitting, existence of overfit solutions within the final population cannot be completely discounted.  It remains a user's responsibility to appropriately and critically compare produced models.  In Galileo's ramp-with-shelf example, identification of such overfit solutions was trivial, but use of a fitness-complexity Pareto front \cite{schmidt2009distilling} or cross-validation could be useful in more complex cases.  

\subsubsection*{Experiment without shelf}
The physical relationship is more complex in Galileo's experiment without the horizontal shelf because the ball enters free fall with an initial downward velocity (and less horizontal velocity).  It is described by the relationship \cite{dickey1995teaching}
\begin{equation}
H = \frac{k_1 D^2}{1 + k_2 D}
\label{eq:galileo_no_shelf}
\end{equation}
where $k_1$ and $k_2$ are dependent upon the geometry of the table and ramp.  The values of the best fit parameters (in a lest-squares sense) for the Galilieo data are $k_1=1.099 \times 10^{-3}$ and $k_2=-1.121 \times 10^{-3}$. The maximum distance traveled occurs at the asymptote of this equation. In other words, the terminal distance is $-1/k_2$ (892 punti in Galileo's example).

\begin{figure}
  \centering
  \begin{subfigure}[b]{0.49\textwidth}
         \centering
         \includegraphics[]{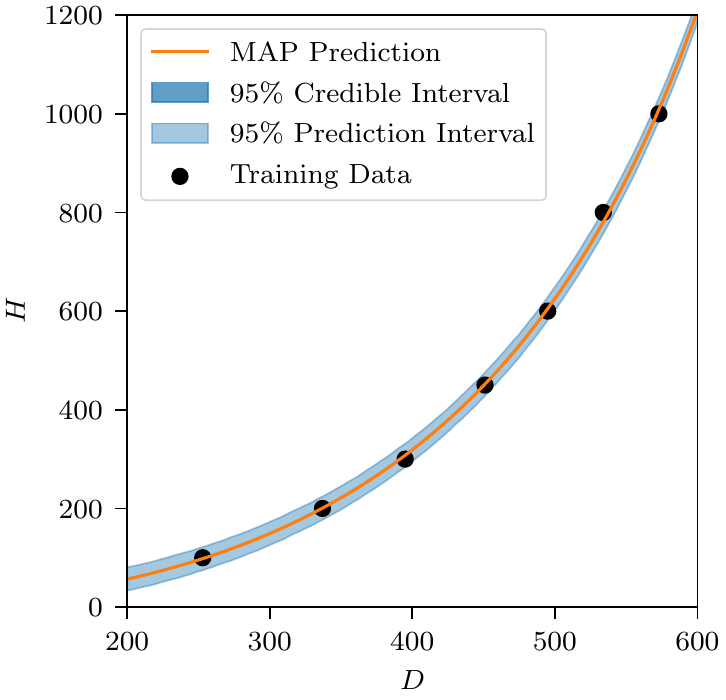}
         \caption{Correct form \eref{eq:galileo_no_shelf_correct_form}}
         \label{fig:galileo_no_shelf_true_form}
     \end{subfigure}
     \hfill
     \begin{subfigure}[b]{0.49\textwidth}
         \centering
         \includegraphics[]{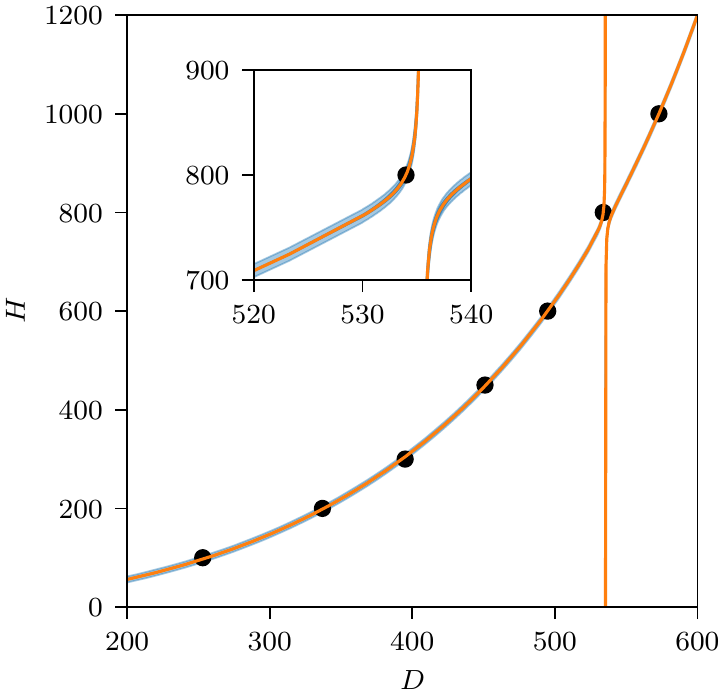}
         \caption{Overfit form \eref{eq:galileo_no_shelf_overfit_form}}
         \label{fig:galileo_no_shelf_overfit_form}
     \end{subfigure}
  \caption{Probabilistic fits produced by Bayesian GPSR for the no-shelf data illustrating the difference between (a) the correct form of the equation and (b) an equation exhibiting overfitting.}
  \label{fig:galileo_diagram}
\end{figure}

\begin{figure}
  \centering
  \begin{subfigure}[b]{0.49\textwidth}
         \centering
         \includegraphics[]{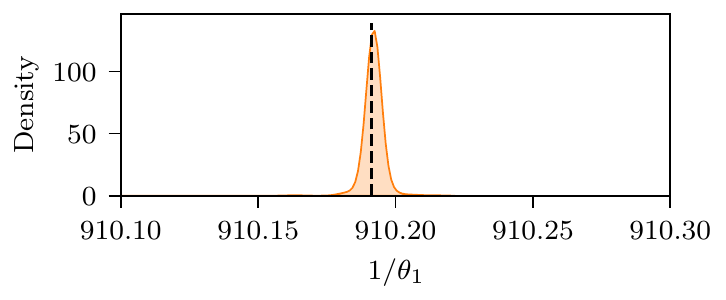}
     \end{subfigure}
     \begin{subfigure}[b]{0.49\textwidth}
         \centering
         \includegraphics[]{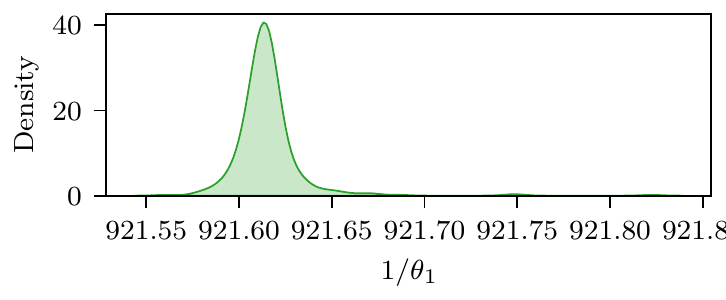}
     \end{subfigure}
  \begin{subfigure}[b]{0.49\textwidth}
         \centering
         \includegraphics[]{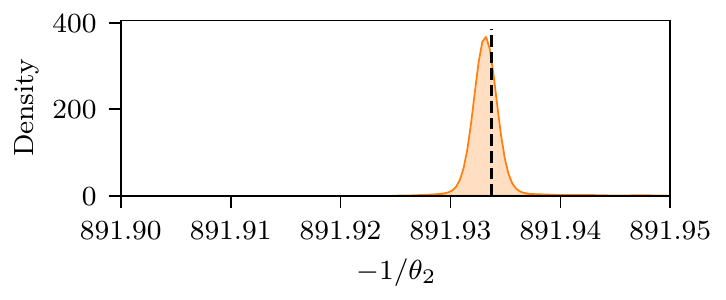}
     \end{subfigure}
     \begin{subfigure}[b]{0.49\textwidth}
         \centering
         \includegraphics[]{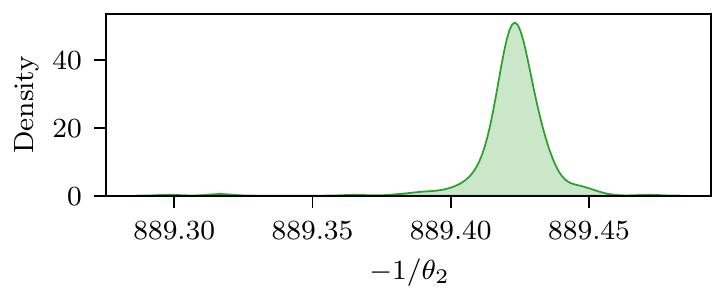}
     \end{subfigure}
  \begin{subfigure}[b]{0.49\textwidth}
         \centering
         \includegraphics[]{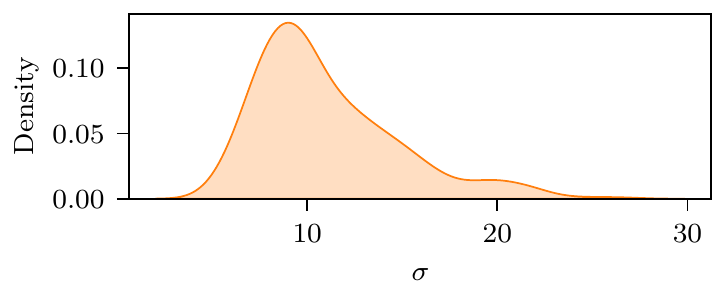}
         \caption{}
         \label{fig:galileo_no_shelf_params_and_noise_true}
     \end{subfigure}
     \begin{subfigure}[b]{0.49\textwidth}
         \centering
         \includegraphics[]{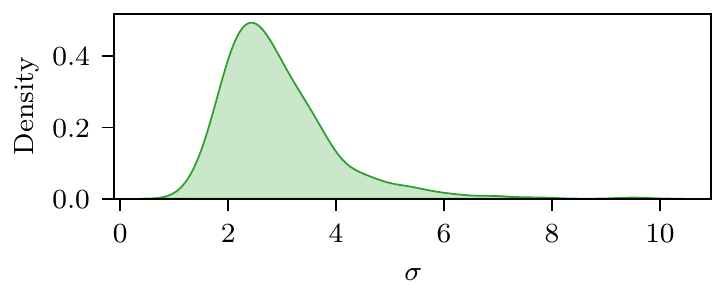}
         \caption{}
         \label{fig:galileo_no_shelf_params_and_noise_overfit}
     \end{subfigure}
  \caption{Estimated probability density functions  of $\theta_1$, $\theta_2$ and  noise standard deviation, $\sigma$ for (a) the correct form of the equation and (b) an equation exhibiting overfitting. Dashed lines indicate the least-squares fits of $\theta_1$ and $\theta_2$}.
  \label{fig:galileo_no_shelf_params_and_noise}
\end{figure}

Bayesian GPSR was performed 11 times with the Galileo ramp-without-shelf data. This data proved to be a much more difficult task for Bayesian GPSR.  The true physical model, e.g.,
\begin{equation}
f(\mathbf{x}, \boldsymbol{\theta}) = \frac{\theta_1 x_0^2}{1 + \theta_2 x_0},
\label{eq:galileo_no_shelf_correct_form}
\end{equation}
was present in the final population in 5 of the 11 repetitions.  A representative result that has this functional form is illustrated in \fref{fig:galileo_no_shelf_true_form}.  The equation does a reasonable job capturing the trend of the data with exception of the datapoint near $D=535$, which is near the upper edge of the 95\% prediction interval.  The parameter estimates of this result are included in \fref{fig:galileo_no_shelf_params_and_noise_true}.  The estimates of the two parameters match well with their least squares estimates.
Of the five Bayesian GPSR repetitions that contained the true physical equation, four contained equations with significantly better description of the data (as measured by FBF).  

One of the equations identified frequently across the repetitions with high FBF is
\begin{equation}
f(\mathbf{x}, \boldsymbol{\theta}) = \frac{x_0^2(\theta_1 +\theta_2 x_0)}{\theta_3 x_0^2 + \theta_4 x_0 + \theta_5}.
\label{eq:galileo_no_shelf_overfit_form_unsimplified}
\end{equation}
A representative result which has this functional form is illustrated in \fref{fig:galileo_no_shelf_overfit_form}.
After reparametrization, \eref{eq:galileo_no_shelf_overfit_form_unsimplified} becomes
\begin{equation}
f(\mathbf{x}, \boldsymbol{\theta}) = \left( \frac{\theta_1 x_0^2}{1+\theta_2 x_0} \right)  \left( \frac{1 +\theta_3 x_0}{1 + \theta_4 x_0} \right) .
\label{eq:galileo_no_shelf_overfit_form}
\end{equation}
The left parenthetical is the true physical equation and the right is an augmentation that allows for the addition of a second asymptote.  In the result shown in \fref{fig:galileo_no_shelf_overfit_form} the values of $\theta_3 \approx \theta_4 \approx -\frac{1}{535.5}$ lead to an augmentation that is approximately $1$  except in the neighborhood of $x_0=535.5$ where there is asymptotic behavior.  The asymptotic behavior allows for the equation to nearly exactly match the datapoint at $D=535$ without significantly modifying the equation elsewhere. Ultimately, this leads to parameter estimates for $\theta_1$ and $\theta_2$ that are similar to their true values and a much smaller estimate of the noise level $\sigma$ (\fref{fig:galileo_no_shelf_params_and_noise_overfit}).

Due to the interpretable nature of Bayesian GPSR some forms of overfitting like \eref{eq:galileo_no_shelf_overfit_form} can be identified and mitigated \emph{a posteriori}; nevertheless, it is worth investigating why this type of overfitting occurred.  It has been shown that Galileo's no-shelf dataset is biased relative to the underlying physics \cite{drake1975galileo, drake2003galileo}, likely due to experimental limitations of the era.  With this in mind it is worth testing the validity of the underlying assumption that the noise in the stochastic model (\eref{eq:stochmodel}) is normally distributed.  To this end, the residuals of Galileo's ramp-without-shelf data relative to the least-squares fit of the true physics were calculated and subsequently used in a Shapiro-Wilk \cite{shaphiro1965analysis} normality test.  The test showed that the residuals are not normally distributed with power $\alpha=0.1$ and $p$ value $0.06$.  On the other hand, the residuals with respect to the overfit \eref{eq:galileo_no_shelf_overfit_form} become more normally distributed ($p$ value of $0.18$).

The lack of normality of the noise in Galileo's no-shelf experiment is investigated further by producing new synthetic datasets where the noise is normally distributed.  To create these datasets, the least-squares fit of the true physics, \eref{eq:galileo_no_shelf}, is used as the data generating function.  For each dataset, $N=7$ realizations of $D$ are drawn from a uniform distribution, $D_i \sim \textrm{Uniform}(0, 800)$, and added to samples from a zero-mean noise distribution with $\sigma$ equal to the standard deviation of the residuals (i.e., approximating the same level of noise as Galileo's experiments).  Six datasets were constructed and used for Bayesian GPSR with hyperparameters identical to those used above.  In all six cases, the true physical equation existed in the final population with the best (or approximately equal to the best) fitness.  In most cases, the true physical form was also the most frequently occuring equation in the population.  Given these results, it can be concluded that the idiosyncrasies of Galileo's no-shelf data is at least partially responsible for the tendency to overfit.  Future efforts could be focused on the incorporation or generation of non-Gaussian noise models to more directly address the issue.

%


\section{Conclusion}
A new Bayesian framework for genetic-programming-based symbolic regression (GPSR) was developed. For each equation in the population, Bayesian inference was used to estimate probability density functions of the unknown constants given the available data. This automatic quantification of uncertainty meant that any equation could be used to make probabilistic predictions using, for example, Monte Carlo simulation. As a byproduct of this process, the normalized marginal likelihood of the fractional Bayes' factor was computed and used to represent the equation fitness. Coupled with the probabilistic crowding algorithm, this new fitness metric enables Bayesian model selection during population evolution where the replacement probability is rooted in model evidence. The impact of this approach on the equations produced during symbolic regression was studied through both a numerical example and a real-world example.  The proposed method was able to produce interpretable, data-driven models that incorporate uncertainty.  Additionally, the proposed method was shown to increase interpretability, improve robustness to noise, and reduce overfitting when compared to a non-Bayesian GPSR implementation.

A number of open issues are left to future work. First, in the examples presented, an additive Gaussian noise assumption was made. While this assumption could easily be altered depending on the application of interest, the choice, once made, is fixed throughout GPSR. There is potential to allow the GPSR algorithm to modify the likelihood function during evolution to better fit the data. Second, it is common to know \textit{a priori} what parameters should appear in the generated equations (e.g., known physical constants). While not currently implemented, this knowledge could be encoded as subjective priors which would then be available to the GPSR algorithm for inclusion during Bayesian inference. It is noteworthy, in closing, that this paper provides a simple but generic approach to incorporating model evidence in GPSR. Any of the plethora of advancements in the field of uncertainty quantification could be incorporated to enhance the flexibility and applicability of the method.

\bibliographystyle{unsrtnat}
\bibliography{refs}

\begin{thebibliography}{40}
\providecommand{\natexlab}[1]{#1}
\providecommand{\url}[1]{\texttt{#1}}
\expandafter\ifx\csname urlstyle\endcsname\relax
  \providecommand{\doi}[1]{doi: #1}\else
  \providecommand{\doi}{doi: \begingroup \urlstyle{rm}\Url}\fi

\bibitem[Adadi and Berrada(2018)]{adadi2018peeking}
Amina Adadi and Mohammed Berrada.
\newblock Peeking inside the black-box: a survey on explainable artificial
  intelligence (xai).
\newblock \emph{IEEE access}, 6:\penalty0 52138--52160, 2018.

\bibitem[Du et~al.(2019)Du, Liu, and Hu]{du2019techniques}
Mengnan Du, Ninghao Liu, and Xia Hu.
\newblock Techniques for interpretable machine learning.
\newblock \emph{Communications of the ACM}, 63\penalty0 (1):\penalty0 68--77,
  2019.

\bibitem[Bhatt et~al.(2020)Bhatt, Xiang, Sharma, Weller, Taly, Jia, Ghosh,
  Puri, Moura, and Eckersley]{bhatt2020explainable}
Umang Bhatt, Alice Xiang, Shubham Sharma, Adrian Weller, Ankur Taly, Yunhan
  Jia, Joydeep Ghosh, Ruchir Puri, Jos{\'e}~MF Moura, and Peter Eckersley.
\newblock Explainable machine learning in deployment.
\newblock In \emph{Proceedings of the 2020 Conference on Fairness,
  Accountability, and Transparency}, pages 648--657, 2020.

\bibitem[Rudin()]{rudin1811please}
C~Rudin.
\newblock Please stop explaining black box models for high stakes decisions.
  arxiv 2018.
\newblock \emph{arXiv preprint arXiv:1811.10154}.

\bibitem[Rudin and Radin(2019)]{rudin2019we}
Cynthia Rudin and Joanna Radin.
\newblock Why are we using black box models in ai when we don’t need to? a
  lesson from an explainable ai competition.
\newblock \emph{Harvard Data Science Review}, 1\penalty0 (2), 2019.

\bibitem[Schmidt and Lipson(2009)]{schmidt2009distilling}
Michael Schmidt and Hod Lipson.
\newblock Distilling free-form natural laws from experimental data.
\newblock \emph{science}, 324\penalty0 (5923):\penalty0 81--85, 2009.

\bibitem[Brunton et~al.(2016)Brunton, Proctor, and
  Kutz]{brunton2016discovering}
Steven~L Brunton, Joshua~L Proctor, and J~Nathan Kutz.
\newblock Discovering governing equations from data by sparse identification of
  nonlinear dynamical systems.
\newblock \emph{Proceedings of the national academy of sciences}, 113\penalty0
  (15):\penalty0 3932--3937, 2016.

\bibitem[Galioto and Gorodetsky(2020)]{galioto2020bayesian}
Nicholas Galioto and Alex~Arkady Gorodetsky.
\newblock Bayesian system id: Optimal management of parameter, model, and
  measurement uncertainty.
\newblock \emph{Nonlinear Dynamics}, 102\penalty0 (1):\penalty0 241--267, 2020.

\bibitem[Bomarito et~al.(2021)Bomarito, Townsend, Stewart, Esham, Emery, and
  Hochhalter]{bomarito2021development}
GF~Bomarito, TS~Townsend, KM~Stewart, KV~Esham, JM~Emery, and JD~Hochhalter.
\newblock Development of interpretable, data-driven plasticity models with
  symbolic regression.
\newblock \emph{Computers \& Structures}, 252:\penalty0 106557, 2021.

\bibitem[Schmidt and Lipson(2007{\natexlab{a}})]{schmidt2007learning}
Michael~D Schmidt and Hod Lipson.
\newblock Learning noise.
\newblock In \emph{Proceedings of the 9th annual conference on Genetic and
  evolutionary computation}, pages 1680--1685, 2007{\natexlab{a}}.

\bibitem[Hirsh et~al.(2021)Hirsh, Barajas-Solano, and
  Kutz]{hirsh2021sparsifying}
Seth~M Hirsh, David~A Barajas-Solano, and J~Nathan Kutz.
\newblock Sparsifying priors for bayesian uncertainty quantification in model
  discovery.
\newblock \emph{arXiv preprint arXiv:2107.02107}, 2021.

\bibitem[Jin et~al.(2019)Jin, Fu, Kang, Guo, and Guo]{jin2019bayesian}
Ying Jin, Weilin Fu, Jian Kang, Jiadong Guo, and Jian Guo.
\newblock Bayesian symbolic regression.
\newblock \emph{arXiv preprint arXiv:1910.08892}, 2019.

\bibitem[Zhang(2000)]{zhang2000bayesian}
Byong-Tak Zhang.
\newblock Bayesian methods for efficient genetic programming.
\newblock \emph{Genetic Programming and Evolvable Machines}, 1\penalty0
  (3):\penalty0 217--242, 2000.

\bibitem[Koza and Koza(1992)]{koza1992genetic}
John~R Koza and John~R Koza.
\newblock \emph{Genetic programming: on the programming of computers by means
  of natural selection}, volume~1.
\newblock MIT press, 1992.

\bibitem[Worm and Chiu(2013)]{worm2013prioritized}
Tony Worm and Kenneth Chiu.
\newblock Prioritized grammar enumeration: symbolic regression by dynamic
  programming.
\newblock In \emph{Proceedings of the 15th annual conference on Genetic and
  evolutionary computation}, pages 1021--1028, 2013.

\bibitem[Udrescu and Tegmark(2020)]{udrescu2020ai}
Silviu-Marian Udrescu and Max Tegmark.
\newblock Ai feynman: A physics-inspired method for symbolic regression.
\newblock \emph{Science Advances}, 6\penalty0 (16):\penalty0 eaay2631, 2020.

\bibitem[Petersen et~al.(2019)Petersen, Larma, Mundhenk, Santiago, Kim, and
  Kim]{petersen2019deep}
Brenden~K Petersen, Mikel~Landajuela Larma, T~Nathan Mundhenk, Claudio~P
  Santiago, Soo~K Kim, and Joanne~T Kim.
\newblock Deep symbolic regression: Recovering mathematical expressions from
  data via risk-seeking policy gradients.
\newblock \emph{arXiv preprint arXiv:1912.04871}, 2019.

\bibitem[Valipour et~al.(2021)Valipour, You, Panju, and
  Ghodsi]{valipour2021symbolicgpt}
Mojtaba Valipour, Bowen You, Maysum Panju, and Ali Ghodsi.
\newblock Symbolicgpt: A generative transformer model for symbolic regression.
\newblock \emph{arXiv preprint arXiv:2106.14131}, 2021.

\bibitem[La~Cava et~al.(2021)La~Cava, Orzechowski, Burlacu, de~Fran{\c{c}}a,
  Virgolin, Jin, Kommenda, and Moore]{la2021contemporary}
William La~Cava, Patryk Orzechowski, Bogdan Burlacu, Fabr{\'\i}cio~Olivetti
  de~Fran{\c{c}}a, Marco Virgolin, Ying Jin, Michael Kommenda, and Jason~H
  Moore.
\newblock Contemporary symbolic regression methods and their relative
  performance.
\newblock \emph{arXiv preprint arXiv:2107.14351}, 2021.

\bibitem[Bomarito(2022)]{bingo}
Geoffrey Bomarito.
\newblock Bingo.
\newblock \url{https://github.com/nasa/bingo}, 2022.

\bibitem[Schmidt and Lipson(2007{\natexlab{b}})]{schmidt2007comparison}
Michael Schmidt and Hod Lipson.
\newblock Comparison of tree and graph encodings as function of problem
  complexity.
\newblock In \emph{Proceedings of the 9th annual conference on Genetic and
  evolutionary computation}, pages 1674--1679, 2007{\natexlab{b}}.

\bibitem[Kommenda et~al.(2013)Kommenda, Kronberger, Winkler, Affenzeller, and
  Wagner]{kommenda2013effects}
Michael Kommenda, Gabriel Kronberger, Stephan Winkler, Michael Affenzeller, and
  Stefan Wagner.
\newblock Effects of constant optimization by nonlinear least squares
  minimization in symbolic regression.
\newblock In \emph{Proceedings of the 15th annual conference companion on
  Genetic and evolutionary computation}, pages 1121--1128, 2013.

\bibitem[De~Melo et~al.(2015)De~Melo, Fowler, and Banzhaf]{de2015evaluating}
Vinicius~Veloso De~Melo, Benjamin Fowler, and Wolfgang Banzhaf.
\newblock Evaluating methods for constant optimization of symbolic regression
  benchmark problems.
\newblock In \emph{2015 Brazilian conference on intelligent systems (BRACIS)},
  pages 25--30. IEEE, 2015.

\bibitem[Mahfoud(1995)]{mahfoud1995niching}
Samir~W Mahfoud.
\newblock \emph{Niching methods for genetic algorithms}.
\newblock PhD thesis, University of Illinois at Urbana-Champaign, 1995.

\bibitem[Schmidt and Lipson(2011)]{schmidt2011age}
Michael Schmidt and Hod Lipson.
\newblock Age-fitness pareto optimization.
\newblock In \emph{Genetic programming theory and practice VIII}, pages
  129--146. Springer, 2011.

\bibitem[Mengshoel and Goldberg(1999)]{mengshoel1999probabilistic}
Ole~J Mengshoel and David~E Goldberg.
\newblock Probabilistic crowding: Deterministic crowding with probabilistic
  replacement.
\newblock 1999.

\bibitem[Galan and Mengshoel(2010)]{galan2010generalized}
Severino~F Galan and Ole~J Mengshoel.
\newblock Generalized crowding for genetic algorithms.
\newblock In \emph{Proceedings of the 12th annual conference on Genetic and
  evolutionary computation}, pages 775--782, 2010.

\bibitem[Murray and Ghahramani(2005)]{murray2005note}
Iain Murray and Zoubin Ghahramani.
\newblock A note on the evidence and bayesian occam’s razor.
\newblock 2005.

\bibitem[Goldstein(2006)]{goldstein2006subjective}
Michael Goldstein.
\newblock Subjective bayesian analysis: principles and practice.
\newblock \emph{Bayesian analysis}, 1\penalty0 (3):\penalty0 403--420, 2006.

\bibitem[Berger(2006)]{berger2006objective}
James Berger.
\newblock The case for objective bayesian analysis.
\newblock \emph{Bayesian analysis}, 1\penalty0 (3):\penalty0 385--402, 2006.

\bibitem[Berger et~al.(2015)Berger, Bernardo, and Sun]{berger2015overall}
James~O Berger, Jose~M Bernardo, and Dongchu Sun.
\newblock Overall objective priors.
\newblock \emph{Bayesian Analysis}, 10\penalty0 (1):\penalty0 189--221, 2015.

\bibitem[O'Hagan(1995)]{ohagan1995fbf}
Anthony O'Hagan.
\newblock Fractional bayes factors for model comparison.
\newblock \emph{Journal of the Royal Statistical Society: Series B
  (Methodological)}, 57\penalty0 (1):\penalty0 99--118, 1995.

\bibitem[Del~Moral et~al.(2006)Del~Moral, Doucet, and Jasra]{del2006sequential}
Pierre Del~Moral, Arnaud Doucet, and Ajay Jasra.
\newblock Sequential monte carlo samplers.
\newblock \emph{Journal of the Royal Statistical Society: Series B (Statistical
  Methodology)}, 68\penalty0 (3):\penalty0 411--436, 2006.

\bibitem[Smith(2013)]{smith2013uncertainty}
Ralph~C Smith.
\newblock \emph{Uncertainty quantification: theory, implementation, and
  applications}, volume~12, page 162.
\newblock Siam, 2013.

\bibitem[Leser(2022)]{smcpy}
Patrick Leser.
\newblock {SMCPy} - {Sequential Monte Carlo with Python}.
\newblock \url{https://github.com/nasa/smcpy}, 2022.

\bibitem[Virtanen et~al.(2020)Virtanen, Gommers, Oliphant, Haberland, Reddy,
  Cournapeau, Burovski, Peterson, Weckesser, Bright, {van der Walt}, Brett,
  Wilson, Millman, Mayorov, Nelson, Jones, Kern, Larson, Carey, Polat, Feng,
  Moore, {VanderPlas}, Laxalde, Perktold, Cimrman, Henriksen, Quintero, Harris,
  Archibald, Ribeiro, Pedregosa, {van Mulbregt}, and {SciPy 1.0
  Contributors}]{2020SciPy-NMeth}
Pauli Virtanen, Ralf Gommers, Travis~E. Oliphant, Matt Haberland, Tyler Reddy,
  David Cournapeau, Evgeni Burovski, Pearu Peterson, Warren Weckesser, Jonathan
  Bright, St{\'e}fan~J. {van der Walt}, Matthew Brett, Joshua Wilson, K.~Jarrod
  Millman, Nikolay Mayorov, Andrew R.~J. Nelson, Eric Jones, Robert Kern, Eric
  Larson, C~J Carey, {\.I}lhan Polat, Yu~Feng, Eric~W. Moore, Jake
  {VanderPlas}, Denis Laxalde, Josef Perktold, Robert Cimrman, Ian Henriksen,
  E.~A. Quintero, Charles~R. Harris, Anne~M. Archibald, Ant{\^o}nio~H. Ribeiro,
  Fabian Pedregosa, Paul {van Mulbregt}, and {SciPy 1.0 Contributors}.
\newblock {{SciPy} 1.0: Fundamental Algorithms for Scientific Computing in
  Python}.
\newblock \emph{Nature Methods}, 17:\penalty0 261--272, 2020.
\newblock \doi{10.1038/s41592-019-0686-2}.

\bibitem[Drake(2003)]{drake2003galileo}
Stillman Drake.
\newblock \emph{Galileo at work: His scientific biography}.
\newblock Courier Corporation, 2003.

\bibitem[Dickey and Arnold(1995)]{dickey1995teaching}
David~A Dickey and J~Tim Arnold.
\newblock Teaching statistics with data of historic significance: Galileo's
  gravity and motion experiments.
\newblock \emph{Journal of Statistics Education}, 3\penalty0 (1), 1995.

\bibitem[Drake and MacLachlan(1975)]{drake1975galileo}
Stillman Drake and James MacLachlan.
\newblock Galileo's discovery of the parabolic trajectory.
\newblock \emph{Scientific American}, 232\penalty0 (3):\penalty0 102--111,
  1975.

\bibitem[Shaphiro and Wilk(1965)]{shaphiro1965analysis}
S~Shaphiro and M~Wilk.
\newblock An analysis of variance test for normality.
\newblock \emph{Biometrika}, 52\penalty0 (3):\penalty0 591--611, 1965.

\end{thebibliography}

\end{document}